 \documentclass[pmlr,twocolumn,10pt]{jmlr} 

\usepackage{chngcntr}
\usepackage{booktabs}
\usepackage[load-configurations=version-1]{siunitx} 

\newcommand{\equal}[1]{{\hypersetup{linkcolor=black}\thanks{#1}}}

\theorembodyfont{\upshape}
\theoremheaderfont{\scshape}
\theorempostheader{:}
\theoremsep{\newline}

\jmlrvolume{LEAVE UNSET}
\jmlryear{2021}
\jmlrsubmitted{LEAVE UNSET}
\jmlrpublished{LEAVE UNSET}
\jmlrworkshop{Machine Learning for Health (ML4H) 2021} 

\title[eCaReNet]{Towards Explainable End-to-End Prostate Cancer Relapse Prediction from H\&E Images Combining Self-Attention Multiple Instance Learning with a Recurrent Neural Network}

\author{%
\Name{Esther Dietrich}$^1$
\Email{esther.dietrich@zmnh.uni-hamburg.de}\\
\Name{Patrick Fuhlert}$^1$ \Email{patrick.fuhlert@zmnh.uni-hamburg.de}\\
\Name{Anne Ernst}$^{1}$ \Email{anne.ernst@zmnh.uni-hamburg.de}\\
\Name{Guido Sauter}$^2$ \Email{g.sauter@uke.de}\\
\Name{Maximilian Lennartz}$^2$ \Email{m.lennartz@uke.de}\\
\Name{H. Siegfried Stiehl}$^3$ \Email{hans-siegfried.stiehl@uni-hamburg.de}\\
\Name{Marina Zimmermann}$^{1,4\equal{Equally contributing last authors}}$ \Email{marina.zimmermann@zmnh.uni-hamburg.de}\\
\Name{Stefan Bonn}$^{1\footnotemark[1]}$ \Email{sbonn@uke.de}
\\ \addr $^1$Institute of Medical Systems Biology, Center for Biomedical AI (bAIome) $^2$Institute of Pathology $^4$III. Department of Medicine -- University Medical Center Hamburg-Eppendorf, Hamburg, Germany \\$^3$Department of Informatics, Universit\"at Hamburg, Hamburg, Germany \\
}

\begin{document}

\maketitle

\begin{abstract}
Clinical decision support for histopathology image data mainly focuses on strongly supervised annotations, which offers intuitive interpretability, but is bound by expert performance. Here, we propose an explainable cancer relapse prediction network (eCaReNet) and show that end-to-end learning without strong annotations offers state-of-the-art performance while interpretability can be included through an attention mechanism. On the use case of prostate cancer survival prediction, using 14,479 images and only relapse times as annotations, we reach a cumulative dynamic AUC of 0.78 on a validation set, being on par with an expert pathologist (and an AUC of 0.77 on a separate test set). Our model is well-calibrated and outputs survival curves as well as a risk score and group per patient. Making use of the attention weights of a multiple instance learning layer, we show that malignant patches have a higher influence on the prediction than benign patches, thus offering an intuitive interpretation of the prediction. Our code is available at www.github.com/imsb-uke/ecarenet.
\end{abstract}


\section{Introduction}
\label{sec:intro}
In recent years deep learning has greatly improved the performance in computer vision tasks for medical applications \citep{Esteva2021}. In particular, decision support systems for cancer treatment in the field of computational pathology are emerging \citep{Abels2019}. 
Many systems rely on physicians' annotations like treatment decisions, manual annotation of tissue regions or patient classification according to a staging system \citep{Bulten2020}. This strong supervision limits the models' performance through subjectivity and thus ambiguity of the ground truth, and emphasizes the need for quantifiable labels that are independent of the physician, such as time to disease-related death or relapse. Difficulties arise as such labels are relatively weak (a single survival time per patient), thus requiring a large dataset for training, and include censored cases as not all patients relapse or die of the disease. A survival model, unlike a classification model, models the patient's survival over time and can include censored patients. 

The aim of this work is to show that high predictive power can be achieved when training end-to-end only with quantitative patient relapse times, while having a majority of censored cases. 
We focus on prostate cancer as a use case, which was the cancer with the second most new cases in men worldwide in 2020 \citep{Ferlay2020}. Instead of predicting Gleason grades to estimate cancer severity \citep{Gleason1974}, which are highly controversial, often revised and suffer from interobserver variability of up to 81\% \citep{Egevad2012}, we use time to biochemical recurrence (BCR) as annotation. This is defined as a significant rise in prostate specific antigen (PSA) level in the blood after prostatectomy. As input, digitized hematoxylin and eosin (H\&E) stained tissue microarray (TMA) spots are used, of which a dataset containing 14,479 images is available.

To the best of our knowledge, we are the first to propose an explainable end-to-end deep learning model to predict BCR over time after prostatectomy from TMA spots. We introduce a novel network based on self-attention \citep{Rymarczyk2020}, attention-based multiple instance learning (MIL, \cite{Ilse2018}) and recurrent neural networks (RNNs, \cite{Rumelhart1987}) for survival prediction, called eCaReNet (explainable cancer relapse prediction network). With an AUC of 0.78 on the validation set (0.77 on the test set) we achieve state-of-the-art results, while assuring calibration. Through evaluation of attention weights of the MIL layer, we further show that our model weights malignant patches higher than benign patches.  In general, our approach is applicable to various cancer and non-cancer histopathology survival prediction problems. 

This work is organized as follows. \sectionref{sec:related_work} gives an overview on related work. In \sectionref{sec:dataset}, the available data is described. The details of our model can be found in \sectionref{sec:methods}, followed by a discussion of the results in \sectionref{sec:results}, including a comparison to benchmark models and a pathologist.

\section{Related Work}
\label{sec:related_work}
Image-based decision support systems often aim at reproducing classification systems used in clinical practice \citep{Bulten2020}. Only after classification \cite{Arvaniti2018} and \cite{Nagpal2019} correlate findings to patient survival. Disadvantages are time-consuming annotations and label quality limited by the annotator. However, this classification allows for an improved interpretability of progression prediction. If only a weak label for a whole image is available, MIL approaches as proposed by \cite{Ilse2018} can be applied to analyze which image regions have the highest influence on a model's prediction. \cite{Couture2018} for example integrate MIL methods in their model for risk of recurrence prediction from image patches, which is also a clinical score. Especially when analyzing whole slide images, MIL approaches are often used. \cite{Yao2020} predict a single risk score per patient with attention-based MIL, while \citep{Campanella2019} use MIL for binary tumor classification. \cite{Lu2021} extend MIL to multi-class classification by implementing multiple attention branches. Image regions relevant for the diagnosis are indicated by high attention weights.

Human performance can be improved upon if disease progression is modeled based on patient outcome directly. A binary classification of whether a patient has a relapse before a certain point in time is often applied, but is diagnostically less conclusive \citep{Duanmu2020, Yamamoto2019}. \cite{Wulczyn2020} treat survival prediction as a multi-class problem with the goal to correctly classify the interval of the event and output a risk score.

To predict relapse probability over time, a survival analysis model can be used. One option is the Cox model \citep{Cox1972}, where the linear part can be replaced by a neural network, as proposed in DeepSurv by \cite{Katzman2018} and its counterpart for images DeepConvSurv by \cite{Zhu2016}. Especially in histopathology, often a complex feature extraction step is applied prior to the Cox model \citep{Yao2020, Tang2019, Zhang2021}. Furthermore, the Cox model is limited by the proportional hazard assumption, which enforces hazard rates to be constant over time. \cite{Xiao2020} and \cite{Vale-Silva2021} avoid the proportional hazard assumption as well as annotation-expensive preprocessing steps by developing end-to-end deep learning models. The latter however also include electronic health record and omics data to improve performance and neither includes an explainability mechanism, treating the model as a black box.  \cite{Ren2019} and \cite{Giunchiglia2018} include recurrent layers to model time dependency, but only use patient electronic health records. A different approach is applied by \cite{Yala2021}, who predict risk of cancer over time from mammography images by converting the prediction into a classification across multiple time points.

We propose a novel framework named eCaReNet for explainable end-to-end relapse prediction by exploiting the advantages of different works.

\section{Dataset}
\label{sec:dataset}
Two datasets were provided by the local pathology department. All images in our datasets show prostate tissue obtained after prostatectomy, during which the patient's prostate is removed. Multiple tissue samples are then taken with a hollow needle, resulting in tissue cores of  0.6 mm diameter each. Arranged in a TMA, multiple samples from multiple patients are stained at once with H$\&$E and digitized afterwards. Small differences, or biases, in staining intensity between TMA blocks arise due to e.g.\ staining times \citep{Parsons2009}. 

The survival dataset (see \tableref{tab:dataset}) comprises 39 digitized TMAs with 129 to 522 images each. For these images, besides the time to BCR and the censoring status, the integrative quantitative (IQ-) Gleason and International Society of Urological Pathology (ISUP) scores \citep{Sauter2018, Egevad2012} of the whole prostate are labeled (for details on Gleason grading see \appendixref{sec:prostate_cancer}). In this context it is important to note that the Gleason scores are based on the whole prostate, while in our dataset the image per patient only represents a small part of prostate tissue.  Since the TMA spot image can only cover a very small part of the prostate, and annotations for individual images are missing, it is possible that a given image is not fully representative of a patient's disease status and outcome. In order to remove the most extreme of those cases, per-image Gleason scores are predicted and compared to the annotated overall Gleason score. Images that are predicted as non-cancerous -- but have a high overall IQ Gleason annotation, a PSA value $>$ 4 $\frac{ng}{ml}$ and a relapse within 2 years -- are removed from the dataset, as these are considered unrepresentative and expected to reduce the model's generalizability. Other discrepancies between the images and relapse times are left unchanged. 

Images from all but one TMA are shuffled and split into training, validation and test sets, stratified by prostate Gleason annotation. One TMA is left out as a separate test set to evaluate model performance on a set with a unique staining bias. \tableref{tab:dataset} summarizes the number of images per dataset split. The distribution of event times for censored and uncensored patients in the training and validation datasets is shown in \figureref{fig:apd_data_distribution}.

We pretrained our model on a second, smaller dataset (Gleason dataset), which also contains TMA spots, but is annotated with image-level Gleason and ISUP scores (see \figureref{fig:apd_model_3_steps}A). It includes 1930 images in the training, 417 in the validation and 419 in the test set. 

For clinical practice, an estimate of the relapse time is of interest prior to prostatectomy. Since tissue samples obtained through needle biopsy are visually similar to post-prostatectomy tissue cores, the dataset can simulate such biopsies. The Gleason labels are annotated following the convention for biopsies.\\
Preprocessing and data augmentation steps are detailed in \appendixref{sec:preprocessing}.

\begin{table}[htbp]
    \centering
    \caption{Overview of number of images for each split in the survival dataset. 80\% of patients are censored. $c\!=\!0$:uncensored, $c\!=\!1$:censored, valid.:validation.}
    \label{tab:dataset}
    \begin{tabular}{c|c|c|c|c}
         &training & valid.  & test  & single TMA test  \\
         \hline
         $c\!=\!0$ & 1965 & 445 & 429 &  36 \\
         $c\!=\!1$ & 8023 & 1698 & 1742 & 141 \\
         \hline
         total & 9988 & 2143 & 2171 & 177 \\
    \end{tabular}
\end{table}

\section{Methods}
\label{sec:methods}

\subsection{Survival prediction}
\label{sec:meth_survival}
The following is derived according to \cite{Kvamme2019}. For a patient with relapse at time $t^*$, the probability to be event-free at time $t$ is modeled via the survival function  
\begin{align}
     S(t) = P(t^* > t). 
\end{align}
The risk of the event to occur in the interval at time $t + \Delta t$, given that it did not occur until time $t$, is expressed with the hazard rate 
\begin{align}
    h(t) = \lim_{\Delta t \rightarrow 0} \frac{P(t \leq t^* < t +\Delta t|t^* \geq t)}{\Delta t}. 
\end{align}
A well-established method for modeling survival functions of individual patients via the hazard rate is the Cox model \citep{Cox1972}. It is limited by the proportional hazard assumption, which assumes that the hazard is constant over time and equal for all patients, therefore not allowing for crossing survival curves. 

In order not to be constrained by this assumption, we directly model the individual hazard functions with a neural network. The time is discretized into time intervals $t_j \in (t_0, ..., t_k)$ and discrete versions of survival and hazard functions are defined as
\begin{align}
    S(t_j) = P(t^* > t_j) \label{eq:s_eq_p}, \\
    h(t_j) = P(t^* = t_j | t^* > t_{j-1}), \\
    S(t_j) = \prod_{k=0}^j(1-h(t_k)). \label{eq:Surv_from_hazard}
\end{align} 
The survival function is a monotonically decreasing function, as can be seen from \autoref{eq:Surv_from_hazard}.  

An important characteristic of survival data is censoring. Not all patients in the dataset experience an event, either because they are lost to follow-up, their event occurs after the end of documentation or they never relapse. These patients are right-censored and here $t^*$ is not the time of the event, but the last observed time without any event.  

\subsection{Model}
\label{sec:model}
As a base model for our proposed survival prediction an InceptionV3 network \citep{Szegedy2015}, pretrained on the ImageNet dataset \citep{Russakovsky2015}, is chosen, while replacing the last layers to perform survival prediction as described below. We chose InceptionV3 as it achieved best results in our experiments. We include two preceding steps (\ref{sec:misup} and \ref{sec:mbin}), before training our survival model eCaReNet in a third step. \figureref{fig:apd_model_3_steps} shows an overview of the presented models and which datasets these are trained on.

\subsubsection{M$_{\text{ISUP}}$}
\label{sec:misup}
In the first step, we additionally pretrain the InceptionV3 model to adapt it to our histopathology domain. Our model M$_{\text{ISUP}}$ takes images from the Gleason dataset as input (\figureref{fig:apd_model_3_steps}A), downsized with bilinear interpolation to $1024\times1024$ pixels, and classifies these into one out of six classes (benign or one of 5 malignant ISUP classes). During training, a cross-entropy loss is used. For training details and results, see \appendixref{apd:isup_class}.

\subsubsection{M$_\text{Bin}$}
\label{sec:mbin}
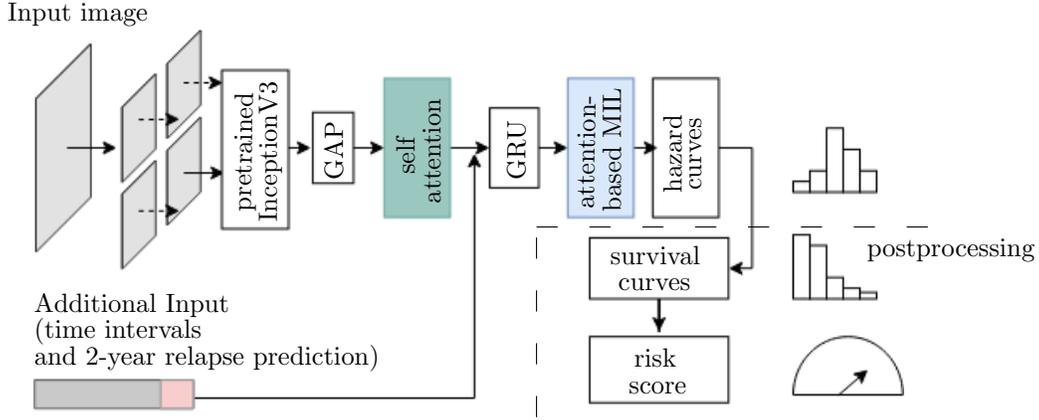
\begin{figure*}
\begin{picture}(100,150)
\def\hshift{50}
\put(\hshift,0){\includegraphics[width=0.7\textwidth]{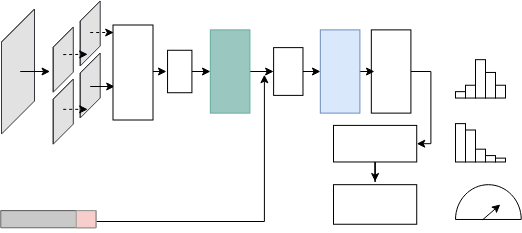}}
\put(\numexpr -10+\hshift,150){Input image}
\put(\numexpr 0+\hshift,40){Additional Input }
\put(\numexpr 0+\hshift,30){(time intervals}
\put(\numexpr 0+\hshift,20){and 2-year relapse prediction)}
\put(\numexpr 75+\hshift,78){\rotatebox{89}{pretrained}}
\put(\numexpr 85+\hshift,74){\rotatebox{90}{InceptionV3}}
\put(\numexpr 110+\hshift,90){\rotatebox{90}{GAP}}
\put(\numexpr 137+\hshift,92){\rotatebox{90}{self}}
\put(\numexpr 147+\hshift,81){\rotatebox{90}{attention}}
\put(\numexpr 179+\hshift,90){\rotatebox{90}{GRU}}
\put(\numexpr 205+\hshift,79){\rotatebox{89}{attention-}}
\put(\numexpr 215+\hshift,78){\rotatebox{90}{based MIL}}
\put(\numexpr 238+\hshift,85){\rotatebox{90}{hazard}}
\put(\numexpr 248+\hshift,86){\rotatebox{90}{curves}}
\put(\numexpr 218+\hshift,58){survival}
\put(\numexpr 221+\hshift,48){curves}
\put(\numexpr 227+\hshift,19){risk}
\put(\numexpr 225+\hshift,9){score}

\put(\numexpr 315+\hshift,61){postprocessing}
\multiput(240,72)(20,0){8}{\line(1,0){8}}
\multiput(240,0)(0,20){4}{\line(0,1){8}}
\end{picture}
\caption{The survival model eCaReNet consists of a pretrained InceptionV3 base model, followed by global average pooling (GAP), a self-attention layer, a recurrent layer with gated recurrent units (GRU) and multiple instance learning (MIL) to combine results per patch. As output, a survival curve as well as a risk score are calculated per patient. The input image needs to be cut into regular patches. As additional information, a prediction whether the relapse occurs in the first two years is used and a time grid is included. The influence of the colored parts is evaluated in \sectionref{sec:results}. }
\label{fig:surv_model_2d}
\end{figure*}

In the second step, a binary classification model M$_{\text{Bin}}$ is used to predict relapse within 2 years on the survival dataset (\figureref{fig:apd_model_3_steps}B). 2 years was chosen, as it lies close to the median (26.8 months) of the relapse times (44\% of relapses earlier than 2 years). For this, we took the model M$_{\text{ISUP}}$ and modified the output to 2 classes. The input image is resized to $1024\times1024$ pixels as in M$_{\text{ISUP}}$ and a cross-entropy loss is applied during training. As opposed to the first step, the prediction per image is saved and used in the third step, which is the survival prediction model eCaReNet, shown in \figureref{fig:surv_model_2d}. 
\subsubsection{eCaReNet}

\label{sec:ecarenet}
Each image of the survival dataset is cut into square, non-overlapping patches as input to eCaReNet (64 patches with $256\times256$ pixels each, see also \sectionref{sec:results}). As this model predicts the hazard over time, one output node per time interval is needed. We chose 28 intervals to cover a time span of 7 years with intervals of 3-months length, covering the 90\% of relapses that occur prior to 7 years. For eCaReNet, only the first 4 inception blocks of M$_{\text{ISUP}}$ are used to reduce overfitting. The following global average pooling layer reduces the dimensionality. Then a self-attention block, as proposed by \cite{Rymarczyk2020}, models the influence of each patch across all other patches. Next, the aforementioned binary classification is concatenated with the output vector of the self-attention layer. This concatenated vector is repeated 28 times to model the discrete time intervals. The current time step is concatenated to each of these vectors. A gated recurrent unit (GRU) layer \citep{Cho2014} models the temporal dependency of the hazard rate in the output, as proposed by \cite{Ren2019}. At the end, an attention-based MIL-layer weights the predictions per patch and outputs a prediction per image, as proposed in \cite{Ilse2018}.  

An individual survival curve per patient is obtained through \autoref{eq:Surv_from_hazard}. Using the normalized area under the survival curve, the patient's overall risk is estimated. Since a large area under the survival curve indicates a low risk $r$ and vice versa, the normalized area is subtracted from one:
\begin{align}
\label{eq:risk}
    r = 1 - \frac{1}{t_k}\sum_{i=1}^{k} S(t_i)\cdot|t_i-t_{i-1}|,
\end{align}
with the last interval $k$ at time $t_k$ (based on the survival time prediction in \cite{Xiao2020}). Since the risk score is a single numerical value between 0 and 1, it eases comparison among patients. 

As proposed by \cite{Kvamme2019}, during training a maximum likelihood loss is optimized. It differs for censored ($c = 1$) and uncensored ($c = 0$) patients with the observed event time $t^*$.
For uncensored patients, the loss $L_u$ can be defined by the known hazard, whereas for censored patients the loss $L_c$ can be defined with the survival function:
\begin{align}
    L_u &= \sum_{c=0} [log(h(t^*)) + \sum_{t_i: t_i<t^*}log(1-h(t_i))], \\
    L_c &= \sum_{c=1} [\sum_{t_i: t_i\leq t^*}log(1-h(t_i))]  \\ 
    &=\sum_{c=1} [log(S(t^*))], \\
    L &= \alpha L_u + (1-\alpha L_c). \label{eq:loss}
\end{align}
Both censored and uncensored patients' losses are linearly combined and equally weighted with $\alpha=0.5$. As labels, the survival and hazard are defined as described in \sectionref{sec:meth_survival}. Since only a discrete event time is known, from \autoref{eq:s_eq_p} it follows that $S(t_j)=1 \ \forall \ t_j < t^*$  and $S(t_j)=0 \ \forall \ t_j \geq t^*$. For the hazard function, $h(t_j)=0 \ \forall \ t_j < t^*$ and $h(t^*) = 1$ hold. The hazard function is not defined for $t_j > t^*$.

\subsection{Metrics}
To evaluate a survival model, both discrimination and calibration need to be considered. Discrimination estimates whether patients are ranked in the correct order, whereas calibration measures how well the predicted survival curves match with the ground truth. 

To evaluate discrimination, we use the cumulative dynamic area under the curve (c/d AUC),
\begin{align}
    \text{c/d AUC}(t) &= P(S_i(t) < S_j(t) | t^*_i \leq t, t^*_j>t) \nonumber \\  & + 0.5 P(S_i(t) = S_j(t) | t^*_i \leq t, t^*_j>t),
\end{align}
which is further integrated over time and weighted by the Kaplan-Meier estimate to account for censored and uncensored patients. Details can be found in \cite{Blanche2019}.
With the c/d AUC the order of the patients' survival probabilities are compared at multiple discrete time points $t$. Censored patients are only comparable to patients with a known survival time that is shorter than the time of censoring. Perfect order results in a measure of 1 \citep{Blanche2019}. To improve readability, we refer to the c/d AUC as AUC in the following. \\ 
In the literature, the concordance, or c-index, is more commonly used \citep{Blanche2019}. This measure also ranges between 0 and 1, with 1 being perfect discrimination. However, the c-index is not a proper scoring rule, meaning that the underlying data generation process does not necessarily give the best score \citep{gerds2021}. 
The Brier score combines calibration and discrimination \citep{Brier1950}, as it measures the mean squared error between the ground truth survival curve and the predicted survival curve. A model that reaches a Brier score below 0.25 is considered to be meaningful \citep{gerds2021}. \\
In an ideal case, the predicted survival curve would be compared to the true survival probability over time, but this cannot be observed. To evaluate how meaningful the resulting survival curves for single patients are, the d-calibration is introduced by \cite{Haider2020}. The idea behind this is to verify that the predicted survival probability at time $t$ matches the true probability of surviving up to time $t$. The d-calibration is calculated by comparing the number of patients that relapse while having a certain predicted survival probability to the expected number. D-calibration is measured with a chi-square test, that needs to pass. For details, see \appendixref{apd:dcal}. \\ 
The Brier score evaluates individual patients' predictions, while the other metrics are only applicable to whole populations. These are thus best used for comparisons between model performances on the same data, not across datasets \citep{gerds2021}. 

\section{Experiments and Results}
\label{sec:results}
All models are trained with the Nadam optimizer \citep{Dozat2016} on the training set and the model with the smallest loss $L$ on the validation set is chosen for evaluation. 5 training runs are performed per model with different random seeds for weight initialization to avoid initialization bias. The models are implemented in Tensorflow 2.1 in Python 3.6. Training is performed on an NVIDIA Quadro RTX 8000 GPU with 48 GB memory. As the focus of this work is on survival prediction, the results for the pretraining on ISUP scores are provided in \appendixref{apd:isup_class}.

\subsection{Benchmark}
\begin{figure}[t]
\floatconts
  {fig:survcurves}
  {\caption{Example survival curves for 3 patients of the survival test dataset, predicted with eCaReNet (blue) and CDOR (orange-red). Both models predict the order of the patients correctly. The survival curves predicted by our model are monotonically decreasing, in contrast to CDOR. $t^*$: time of BCR in months. For eCaReNet, also the predicted risk and risk group are indicated.}}
  {\includegraphics[width=\linewidth]{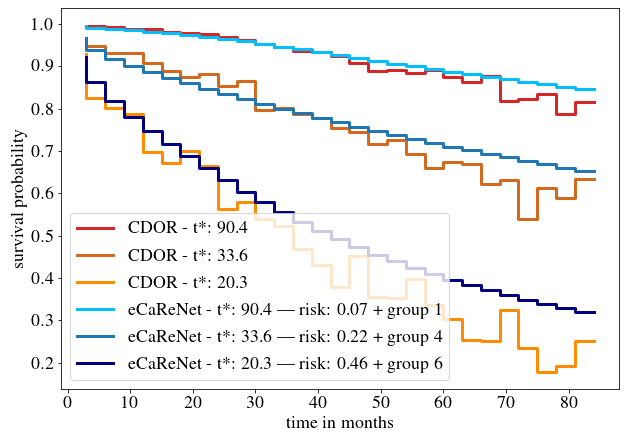}}
\end{figure}
\begin{figure}
    \floatconts
    {fig:risk_stratification}
    {\caption{Kaplan-Meier curves for separate risk groups on the test set. The majority of groups separate well, only the log-rank tests between groups 2/3 as well as 4/5 fail with p-values 0.07 and 0.31 respectively.}}
    {\includegraphics[width=\linewidth]{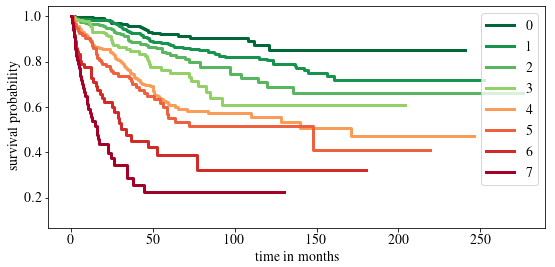}}
\end{figure}
\begin{table*}[htbp]
\floatconts
  {tab:model_benchmark_comparison}%
  {\caption{Benchmark of our proposed approach. Values are the mean of 5 training runs with the standard deviation in parentheses. For d-calibration only failure (f) or pass (p) is indicated.}}%
  {\begin{tabular}{l|llll}
  \toprule
  \bfseries Validation set & \bfseries AUC & \bfseries Brier & \bfseries C-index  & \bfseries D-calibration \\
  \midrule \midrule
  ISUP & \textbf{0.78} & - & \textbf{0.75} & - \\
  \\
  DeepConvSurv \citep{Zhu2016} & 0.69 (0.0207) & 0.305 (0.0146) & 0.65 (0.0173) & f \\
  CDOR \citep{Xiao2020} & 0.77 (0.0089) & 0.111 (0.0014) & 0.73 (0.0046) & f \\
  \textbf{eCaReNet} & \textbf{0.78} (0.0041) & \textbf{0.107} (0.0004) & \textbf{0.75} (0.0016) & \textbf{p} \\
  \midrule \midrule
  \textbf{Test set} 
  \\\midrule \midrule
  ISUP  & \textbf{0.80} & - & \textbf{0.76} & - \\ \\
  DeepConvSurv  &  0.71 (0.0232) & 0.296 (0.0227) & 0.64 (0.0132)& f \\
  CDOR & \textbf{0.78} (0.0005) & 0.110 (0.0001) & 0.73 (0.0003) & f \\
  \textbf{eCaReNet} & 0.77 (0.0048) & \textbf{0.109} (0.0006) & \textbf{0.74} (0.0037) & \textbf{p} \\
  \bottomrule
  \end{tabular}}
\end{table*}
As benchmark, we compare eCaReNet to architectures and loss functions proposed in the literature as well as to an expert pathologist's annotations. Results are summarized in \tableref{tab:model_benchmark_comparison}, where higher AUC and c-index, but lower Brier score indicate better model performance. For d-calibration, only pass or failure of the chi-square test is indicated. 
We start by comparing eCaReNet to two models proposed in the literature. First, we retrain our pretrained M$_{\text{ISUP}}$ (see \sectionref{sec:misup}) with the Cox loss and output as proposed in DeepConvSurv by \cite{Zhu2016}. To do so, the output needs to be reduced to only one node. That model reaches an AUC of 0.69 (c-index of 0.65) on the validation and 0.71 (0.64) on the test set. The test for d-calibration fails and also the Brier score of 0.305 (0.296 on the test set) indicates a non-calibrated model. As second comparison, we train M$_{\text{ISUP}}$ with the output structure and loss proposed in censoring-aware deep ordinal regression (CDOR) by \cite{Xiao2020}. That model reaches an AUC of 0.77 on the validation and 0.78 on the test set and a c-index of 0.73 for both sets, but also fails in terms of calibration. Furthermore, the resulting survival curves are not monotonically decreasing, therefore biologically unreasonable (see \figureref{fig:survcurves}). 
\\
Compared to both previously described models, eCaReNet shows the best performance for all measures on the validation set (AUC 0.78, Brier score 0.107, c-index 0.75) and passes the chi-square test for d-calibration. On the test set it also obtains the best Brier score (0.109) and c-index (0.74) and passes the chi-square test for d-calibration. CDOR performs best on the test set's AUC. In contrast to CDOR, eCaReNet outputs monotonically decreasing survival functions (see \figureref{fig:survcurves}). 
\\
In addition, we assign individual patients to 8 risk groups, to enable a relative ranking as detailed in \appendixref{apd:risk_strat}. Risk groups further allow the evaluation of Kaplan-Meier curves \citep{Kaplan1958} with a log-rank test, which is common in survival analysis \citep{Li2015}. Kaplan-Meier curves are calculated for the risk groups on the training, validation and test datasets. Overall, we can show that the risk groups stratify well on all sets. The results for the test set are shown in \figureref{fig:risk_stratification}, where five out of seven log-rank tests pass ($p<0.05$). 
\\
Furthermore, we compare eCaReNet to annotations of an expert pathologist. In clinical practice, pathologists do not estimate relapse times for patients directly, but assign a Gleason score. We compare eCaReNet's discrimination power to the assigned ISUP scores, since a higher ISUP score corresponds to an increased risk of relapse. eCaReNet reaches on par performance in terms of AUC and c-index with the pathologist's annotations on the validation set (AUC 0.78 and c-index 0.75). Only on the test set, the ISUP annotation shows higher AUC and c-index. In contrast to our model that uses a single TMA spot image per patient, for the ISUP annotation the whole prostate tissue was available, giving Gleason-based survival prediction an advantage over model-based prediction. 

\begin{table*}[htbp]
\floatconts
  {tab:model_ablation_comparison}%
  {\caption{Comparison of model adaptations. Values are the mean of 5 training runs with the standard deviation in parentheses for the validation (Valid.) and test sets. When models M$_{\text{ISUP}}$ or M$_{\text{Bin}}$, or MIL or self-attention (sa) layers are included, it is indicated with a dot ($\bullet$). Best results are marked in bold. MIL=multiple instance learning, Bin=including binary relapse prediction from M$_{\text{Bin}}$. For d-calibration (D-cal.) only failure (f) or pass (p) is indicated.}}
  {\begin{tabular}{l|llll|llll}
  \toprule
  \bfseries Valid. set & \bfseries M$_{\text{ISUP}}$&\bfseries MIL& \bfseries M$_{\text{Bin}}$& \bfseries sa& \bfseries AUC & \bfseries Brier & \bfseries C-index  & \bfseries D-cal. \\
  \midrule \midrule
  M$_{\text{base}}$ &&&&& 0.74 (0.0042) & 0.116 (0.0038) & 0.72 (0.0008) & \textbf{p} \\
  M$_{\text{pretr}}$ &$\bullet$&&&&  0.76 (0.0018) & 0.109 (0.0005) & 0.73 (0.0023) & \textbf{p} \\
  M$_{\text{MIL}}$&$\bullet$&$\bullet$&&&  0.76 (0.0004) & 0.109 (0.0000) & 0.74 (0.0000) & \textbf{p} \\
  M$_{\text{MIL-Bin}}$&$\bullet$&$\bullet$&$\bullet$&&  0.77 (0.0012) & \textbf{0.107} (0.0003) & 0.74 (0.0026) & \textbf{p} \\
  \textbf{eCaReNet} &$\bullet$&$\bullet$&$\bullet$&$\bullet$&  \textbf{0.78} (0.0041) & \textbf{0.107} (0.0004) & \textbf{0.75} (0.0016) & \textbf{p} \\
  \midrule \midrule
  \textbf{Test set}
  \\ \midrule \midrule
  M$_{\text{base}}$ &&&&&  0.74 (0.0054) & 0.115 (0.0007) & 0.71 (0.0031) & \textbf{p} \\
  M$_{\text{pretr}}$ &$\bullet$&&&&  0.76 (0.0031) & 0.110 (0.0004) & 0.73 (0.0018) & \textbf{p}  \\
  M$_{\text{MIL}}$ &$\bullet$&$\bullet$&&&  0.76 (0.0002) & 0.110 (0.0000) & \textbf{0.74} (0.0003) & \textbf{p}  \\
  M$_{\text{MIL-Bin}}$ &$\bullet$&$\bullet$&$\bullet$&&  \textbf{0.77} (0.0011) & \textbf{0.109} (0.0003) & \textbf{0.74} (0.0022) & \textbf{p} \\
  \textbf{eCaReNet} &$\bullet$&$\bullet$&$\bullet$&$\bullet$&  \textbf{0.77} (0.0048) & \textbf{0.109} (0.0006) & \textbf{0.74} (0.0037) & \textbf{p} \\
  \bottomrule
  \end{tabular}}
\end{table*}

\subsection{Comparison of model adaptations}
In the following, eCaReNet (see \figureref{fig:surv_model_2d}) is adapted to evaluate which parts contribute most to model discrimination power and calibration (see \tableref{tab:model_ablation_comparison}). As base model M$_{\text{base}}$, the first 4 blocks of an InceptionV3 model, pretrained on the ImageNet dataset, are extended with a GRU layer for survival prediction. The following adaptations are included gradually. As first adaptation (M$_{\text{pretr}}$), a retraining of the InceptionV3 on Gleason classes as described in M$_{\text{ISUP}}$ is applied (see \sectionref{sec:misup}). The next adaptation is model M$_{\text{MIL}}$, which processes image patches. Here, an attention-based MIL layer is added to the previous model (blue part in \figureref{fig:surv_model_2d}). For model M$_{\text{MIL-Bin}}$, the binary relapse prediction in M$_{\text{Bin}}$ from \sectionref{sec:model} is added (red part in \figureref{fig:surv_model_2d}). The last evaluated model is eCaReNet, where additionally a self-attention layer is included (green part in \figureref{fig:surv_model_2d}), to account for inter-patch influences. For model M$_{\text{MIL}}$ and M$_{\text{MIL-Bin}}$, the experiments showed best results using 16 patches of size $512\times512$ pixels, whereas 64 patches with $256\times256$ pixels each lead to best results when including self-attention.\\
All results are summarized in \tableref{tab:model_ablation_comparison}. It can be seen that the pretraining on histopathology images has a positive effect on all metrics. Adding MIL further improves the discrimination on the validation dataset. Best results are achieved when adding the 2-year relapse prediction and self-attention, reaching performance rivaling that of expert pathologists (AUC validation set: 0.78, test set: 0.77). However, the model with self-attention shows a slightly higher variance in the results than M$_{\text{MIL-Bin}}$. It is concluded that the inter-dependence of patches does not add much additional information to the prediction, as both model versions with and without self-attention show similar performance on the test set. The Brier score is similar for all models and best also for the variants that include a binary relapse prediction. For all our models, the d-calibration chi-square test passes, assuring calibration. Furthermore, the models generalize well, as there is only a slight performance drop when evaluating on the test set. Evaluation of the results on the separate test set, only containing a single TMA, also results in AUC scores of 0.74-0.76 for all adaptations. 

\subsection{Evaluation of attention weights}
\label{sec:attention}
\begin{figure}[htb]
\floatconts
  {fig:boxplots}
  {\caption{In the example images, lighter patches indicate higher attention weights. Boxplots (b) and (d) show results over all example images per experiment. (a) An example image of the first experiment, with the lower part being benign and the upper part being malignant. (c) An example image with the tumor region marked in black. Patches with highest attention weights all lie within the tumor area. For both experiments, the attention weights for malignant patches are significantly higher than for benign patches ($\ast:p<0.05$, $\ast\ast:p<0.01$). }}
  {\subfigure[]{\label{fig:fake_attention_example_img} \includegraphics[width=0.53\linewidth]{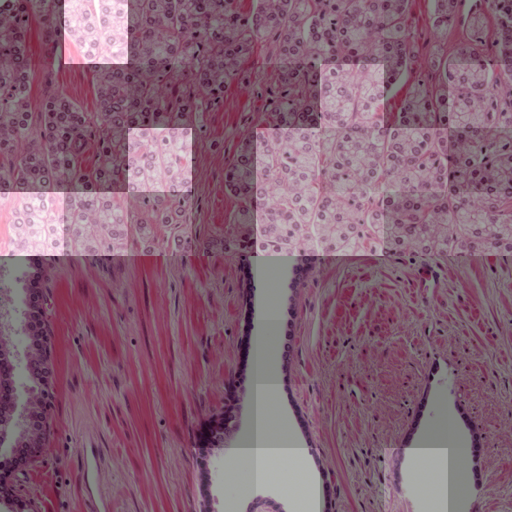}}
  \subfigure[]{\label{fig:fake_attention_example_box} \includegraphics[width=0.4\linewidth]{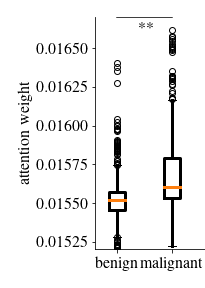}}
  \subfigure[]{\label{fig:annotation_example_img} \includegraphics[width=0.55\linewidth]{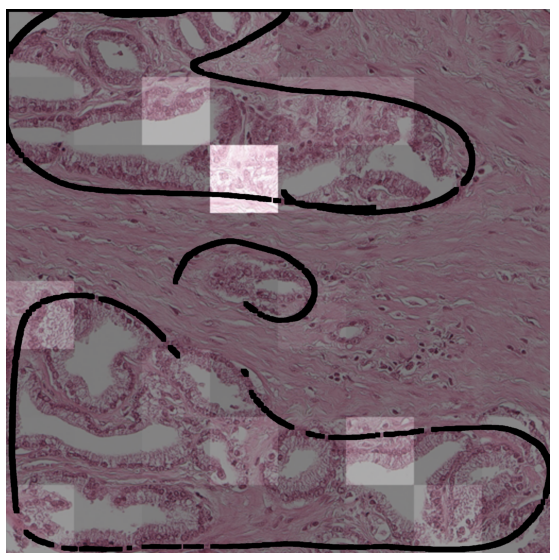}}
  \subfigure[]{\label{fig:annotation_example_box} \includegraphics[width=0.4\linewidth]{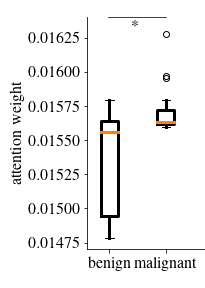}}
  }
\end{figure}

To apply a model in clinical practice, an accurate performance on test data is not sufficient. Physicians can only benefit from a support system if the decisions can be explained and interpreted, with the terms `explainability' or `interpretability' having many different and non-standardized meanings in the literature \citep{BarredoArrieta2020}. In this paper, we include explainability by computing the attention weights of the MIL layer and showing which image regions have the highest influence on the prediction. It is expected that malignant patches show higher attention weights than patches with benign tissue. \\
In a first analysis we use the Gleason dataset to create an artificial dataset for which the annotation per image patch is known. Each image in this dataset combines one image showing benign tissue and one image with malignant Gleason grade 5 tissue by stitching half of each together (see example in \figureref{fig:fake_attention_example_img}). For each image, the attention weights per patch are extracted from the MIL layer of eCaReNet. In the example it can be seen that the upper, malignant part, receives the highest attention weights, while in the benign tissue only relatively bright regions are highlighted. This may be because white regions correspond to glands, which are an important structure to distinguish benign from malignant tissue (see also \figureref{fig:apd_att}). A boxplot of the attention weights of all 12 example images is shown in \figureref{fig:fake_attention_example_box}. The attention weights for malignant patches are significantly higher than for benign patches. The original images that were stitched together are neither part of the training nor of the validation or test sets and give an unbiased estimate of importance. \\
Another experiment was conducted on the survival dataset. From each TMA, one image was randomly chosen from both the validation and test sets of the survival dataset, while maintaining the overall data distribution with respect to the ISUP grades, relapse time and censoring status. An expert pathologist marked tumor regions in each image, enabling us to compare this to the attention weights per patch. A patch is counted as tumorous if 66\% of it lie within the marked tumor region. \figureref{fig:annotation_example_img} shows that all highlighted patches lie within the tumor area, however not all patches in the tumor area receive a high attention weight. \figureref{fig:annotation_example_box} shows the results on all images showing tumor tissue drawn from the test set. Patches marked as tumor show on average higher attention weights than non-tumor image patches.  \\
Overall, both experiments provide strong evidence that eCaReNet focuses on tumor regions, thus human interpretable explanations are provided.

\section{Conclusion}
We developed an end-to-end deep learning model for predicting prostate cancer patients' time to relapse using only images as data source. By directly utilizing time to relapse as ground truth, we could show that detailed annotations are not necessary for training, but are useful for pretraining on a small dataset. eCaReNet reaches the same AUC on the validation set that can be reached with ISUP scores, annotated by an expert pathologist on the whole prostate, while our model only uses a single image per patient. 

By including explainability in our model, we tackle a major drawback of end-to-end systems. With an attention module, we open up the black box and showed in two experiments that the model weights malignant patches significantly higher than benign patches. \\
Since eCaReNet only requires pairs of histopathology images and physician-independent labels, it is generalizable and can be applied to other use cases and end points, like time to disease-related death. \\
Future work includes a more detailed analysis of the model in terms of explainability. 
Further improvements are expected when including multiple images per patient or adding additional information, e.g.\ about the patient's age, PSA value or family history.  
\acks{The authors would like to thank Pierre Machart for his contributions especially during planning and creation of this project.}

\bibliography{literature}

\clearpage 

\appendix

\counterwithin{figure}{section}

\section{Dataset and preprocessing}
\subsection{Prostate cancer grading}
\label{sec:prostate_cancer}
If prostate cancer is suspected, the amount of tumor and its grade are first estimated with a biopsy \citep{Grignon2018}. Among different treatment options, a prostatectomy might be chosen. For both biopsy and prostatectomy, the tissue is graded according to the Gleason grading system \citep{Gleason1974}. The tumor is stratified into five Gleason patterns. The Gleason score is defined as the sum of two patterns (in biopsy the most common and the worst, in prostatectomy the two most common patterns). As there is controversy about the grading system, the International Society of Urological Pathology (ISUP) decided on a scoring system that combines different Gleason pattern combinations into five groups \citep{Egevad2012}. However, there is no consensus yet on how to include possible tertiary patterns and also the percentages of the Gleason patterns in the tumor are neglected. Therefore, \cite{Sauter2018} introduced a more differentiated score, the Integrated Quantitative (IQ) Gleason score. In this work, we focus on the ISUP scores, as they are used in most other studies. 

\subsection{Dataset distribution}
\label{apd:dataset_distribution}
\begin{figure}
    \centering
    \includegraphics[width=\linewidth]{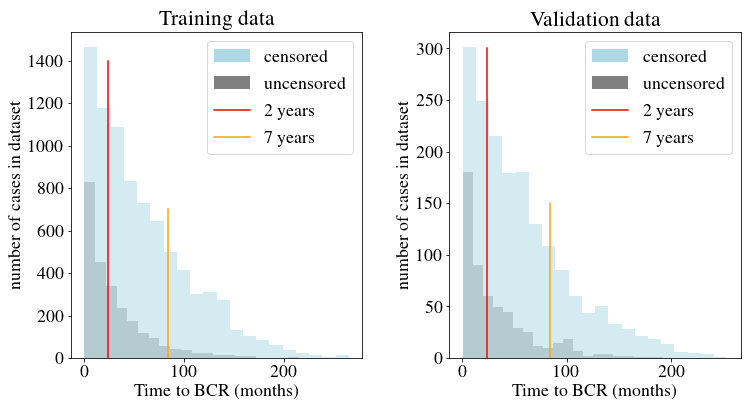}
    \caption{Distribution of censored and uncensored cases in training and validation sets. The red and orange lines indicate 24 and 84 months after prostatectomy.}
    \label{fig:apd_data_distribution}
\end{figure}

In total, our survival dataset contains 17,230 images with prostate tissue, of which 60 images that either contain little to no tissue or are of poor quality (e.g.\ artifacts in the image) are omitted. Additionally, 3,624 patients with unknown relapse time or censoring status are excluded. 709 patients fall under the filtering criterion described in \sectionref{sec:dataset}. Since multiple exclusion criteria may apply to one patient, 14,479 patients remain in the final survival dataset. The dataset distribution is shown in \figureref{fig:apd_data_distribution} for the training and validation sets. 90\% of uncensored events occur prior to 7 years, 44\% prior to 2 years. 

\subsection{Preprocessing}
\label{sec:preprocessing}
The images are all square but of different sizes ($2490\times2490$ to $3181\times3181$ pixels), cut out from  digitized TMA images. Therefore, they consist of circular tissue on white background. Since the white background does not include any information, only a center square of each circular spot is used, which results in images of size $2048\times2048$. The steps are shown in \figureref{fig:meth_dataset_preprocessing_cut_square}. Using the OpenCV package for Python \citep{opencv}, the RGB image \ref{a} is first converted to grayscale \ref{b} and binarized with Otsu-thresholding \ref{c} \citep{Otsu1979}. Then, an ellipse is fitted \ref{d} to use its center as center point for the resulting square \ref{e}-\ref{f}, as not all tissue spots are perfectly round. It is assumed that the information loss at the margins by excluding some tissue is negligible compared to the gain in the foreground to background ratio. 

During model training, data augmentation is applied. As data augmentation methods, the images are randomly flipped and rotated. For the survival model, images are further cut into regular, non-overlapping tiles. The continuous time to BCR label is converted to a binary vector of length 28 for 28 time intervals $t_j$. It has value $1$ for $t_j<t^*$ and $0$ for $t_j\geq t^*$.
\def\mywidth{0.06}
\begin{figure}[ht]
    \centering    
    \subfigure[]{\label{a}\includegraphics[width=\mywidth\textwidth, trim={4cm 1cm 4cm 1cm}, clip]{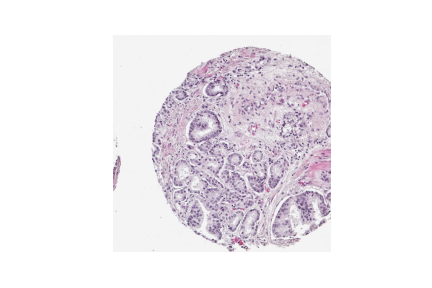}}
    \subfigure[]{\label{b}
    \includegraphics[width=\mywidth\textwidth, trim={4cm 1cm 4cm 1cm}, clip]{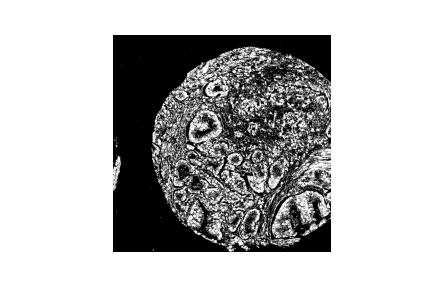}}
    \subfigure[]{\label{c}\includegraphics[width=\mywidth\textwidth, trim={4cm 1cm 4cm 1cm}, clip]{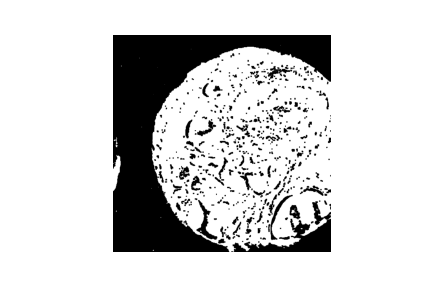}   }
    \subfigure[]{\label{d}\includegraphics[width=\mywidth\textwidth, trim={4cm 1cm 4cm 1cm}, clip]{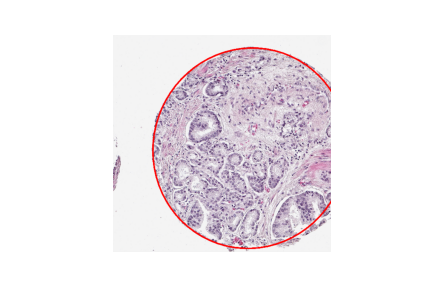}    }
    \subfigure[]{\label{e}\includegraphics[width=\mywidth\textwidth, trim={4cm 1cm 4cm 1cm}, clip]{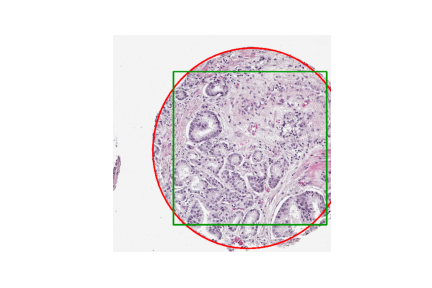}}
    \subfigure[]{\label{f}\includegraphics[width=\mywidth\textwidth, trim={4cm 1cm 4cm 1cm}, clip]{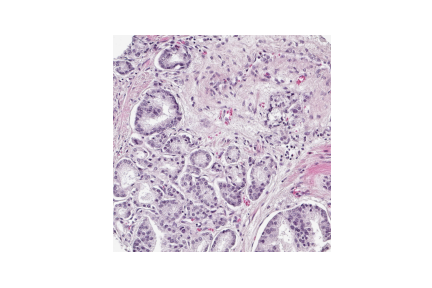}     }

    \caption{Preprocessing steps to cut center pieces of the image and remove most of the white background on an example. Starting with the original image (a), it is converted to grayscale (b), then Otsu-thresholding is applied (c). Next, an ellipse is fitted to the tissue spot, here projected back to the RGB image (d) and a center square is cut (e) - (f).}
    \label{fig:meth_dataset_preprocessing_cut_square}
\end{figure}

\section{ISUP classification}\label{apd:isup_class}
For the ISUP classification in model M$_{\text{ISUP}}$, the ISUP score labels are encoded as an ordinal regression, to account for the fact that e.g.\ classes 2 and 3 are closer than classes 2 and 5. Labels are in the form $y = [l_i]$ for $i=0...4$ with $l_i = 1$ if $i<c$ else $0$ for each class c. For example, class 2 is encoded as $[1,1,0,0,0]$. The model's output $o$ is converted back to class label by summing all output values $p = \sum_i o_i$ and rounding the result.
During training, the cross-entropy loss, 
\begin{align}
    l = \sum_C y_{c} \log (o_{c}),
\end{align}
introduced by \cite{Rubinstein1999}, is optimized using $C$ classes. 

The ISUP classification was trained on 1863 images with 402 validation images of the Gleason dataset (see \figureref{fig:apd_model_3_steps}A). The kappa score on the validation set is 0.85. The confusion matrix is shown in \figureref{fig:apd_confmatrix}.
\begin{figure}[hb]
\floatconts
    {fig:apd_confmatrix}
    {\caption{Confusion matrix for ISUP classification. The axes show the ISUP scores as well as the corresponding possible Gleason grade combinations. Most class confusions are between neighboring classes.}}
    {\includegraphics[width=\linewidth]{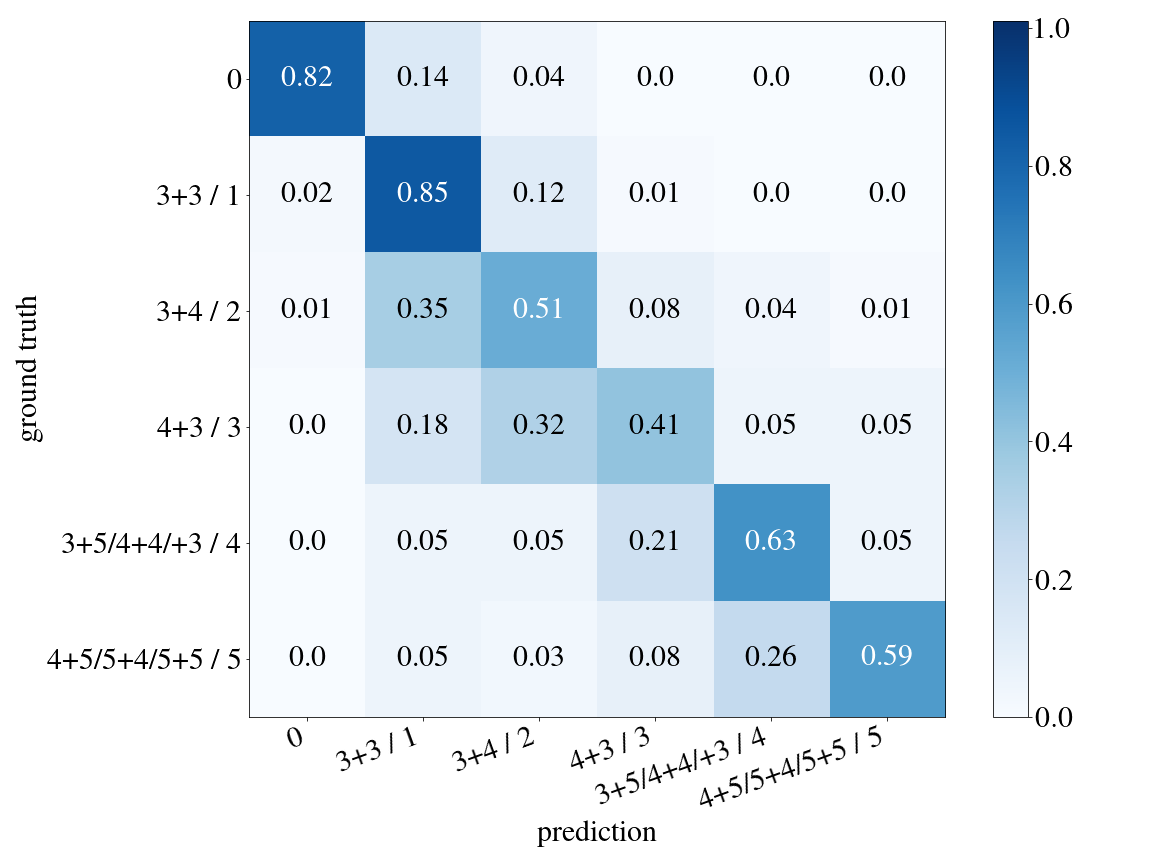}}
\end{figure}

\section{Model}
\label{apd:model}
\begin{figure*}[bp]
\begin{picture}(400,300)
\put(105,0){\includegraphics[width=0.7\textwidth]{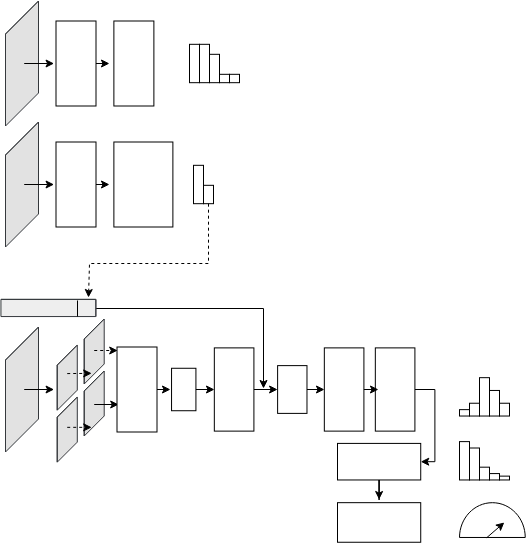}}
\put(105,168){Additional}
\put(120,157){input}
\put(105,358){Input}
\put(105,348){image}

\put(30,350){\rotatebox{90}{\textbf{Dataset}}}
\put(45,350){\rotatebox{90}{\textbf{Loss}}}
\put(60,350){\rotatebox{90}{\textbf{Annotation}}}
\put(85,350){\rotatebox{90}{\textbf{Model}}}
\put(30,290){\rotatebox{90}{Gleason}}
\put(45,280){\rotatebox{90}{cross-entropy}}
\put(60,285){\rotatebox{90}{ISUP score}}
\put(30,210){\rotatebox{90}{survival}}
\put(45,200){\rotatebox{90}{cross-entropy}}
\put(60,190){\rotatebox{90}{relapse $<$ 2 years}}
\put(30,70){\rotatebox{90}{survival}}
\put(45,50){\rotatebox{90}{maximum-likelihood}}
\put(60,40){\rotatebox{90}{relapse time + censoring}}
\put(75,380){\line(0,-1){350}}
\put(25,345){\line(1,0){75}}

\put(0, 301){A}
\put(0, 222){B}
\put(0, 83){C}

\put(85, 280){\rotatebox{90}{M$_{\text{ISUP}}$}}
\put(85, 210){\rotatebox{90}{M$_{\text{Bin}}$}}
\put(85, 70){\rotatebox{90}{eCaReNet}}

\put(180,75){\rotatebox{90}{pretrained}}
\put(190,71.5){\rotatebox{90}{InceptionV3}}
\put(216.5,87){\rotatebox{90}{GAP}}
\put(244,88){\rotatebox{90}{self-}}
\put(252,79){\rotatebox{90}{attention}}
\put(285,87){\rotatebox{90}{GRU}}
\put(310,78){\rotatebox{90}{attention}}
\put(320,75){\rotatebox{90}{based MIL}}
\put(343,83){\rotatebox{90}{hazard}}
\put(353,84){\rotatebox{90}{curves}}
\put(324,55){survival}
\put(328,45){curves}
\put(332,17){risk}
\put(330,7){score}

\put(141,202){\rotatebox{90}{pretrained}}
\put(151,198.5){\rotatebox{90}{InceptionV3}}
\put(179,212){\rotatebox{90}{Binary}}
\put(189,212){\rotatebox{90}{relapse}}
\put(199,205){\rotatebox{90}{prediction}}

\put(141,280){\rotatebox{90}{pretrained}}
\put(151,275){\rotatebox{90}{InceptionV3}}
\put(179,291){\rotatebox{90}{ISUP}}
\put(189,274){\rotatebox{90}{classification}}
\end{picture}
\caption{Overview of the complete model, including the three steps M$_{\text{ISUP}}$, M$_{\text{Bin}}$ and eCaReNet. On the left, the dataset, the loss and the annotation for training are indicated. ISUP: International Society of Urological Pathology, Bin: Binary, GAP: Global Average Pooling, GRU: Gated Recurrent Unit, MIL: Multiple Instance Learning.}
\label{fig:apd_model_3_steps}
\end{figure*}
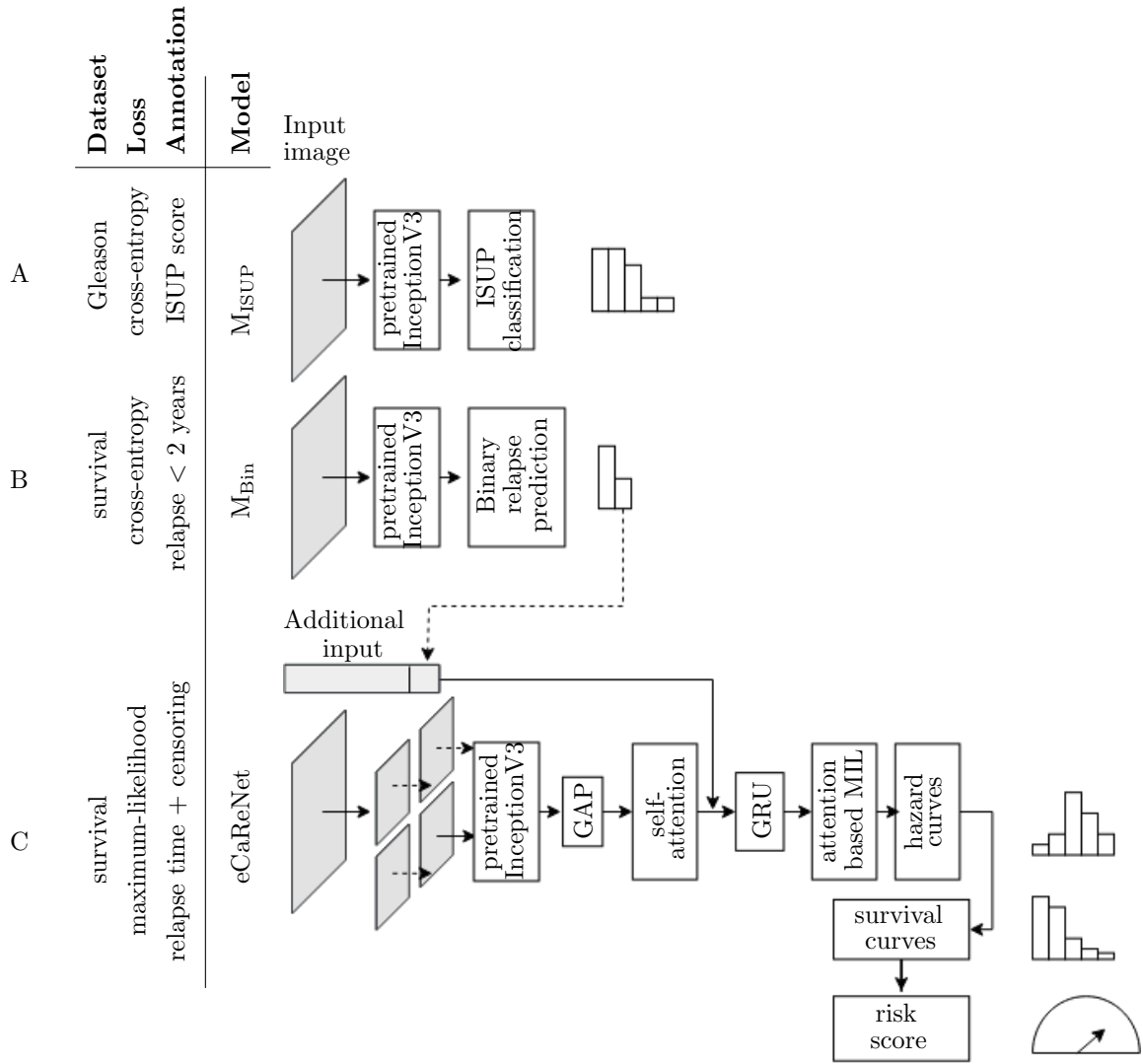

The complete model eCaReNet is shown in \figureref{fig:apd_model_3_steps} including the pretraining model M$_{\text{ISUP}}$ and M$_{\text{Bin}}$.

\subsection{Model results}
The d-calibration for eCaReNet is shown in \figureref{fig:apd_result_dcal}. The distribution's uniformity has been confirmed with a chi-square test.
The AUC of eCaReNet can also be evaluated over time, as shown in \figureref{fig:auc_time}.

\begin{figure}
\floatconts
{fig:apd_result_dcal}
{\caption{Resulting d-calibration plot for eCaReNet. }}
{\includegraphics[width=\linewidth]{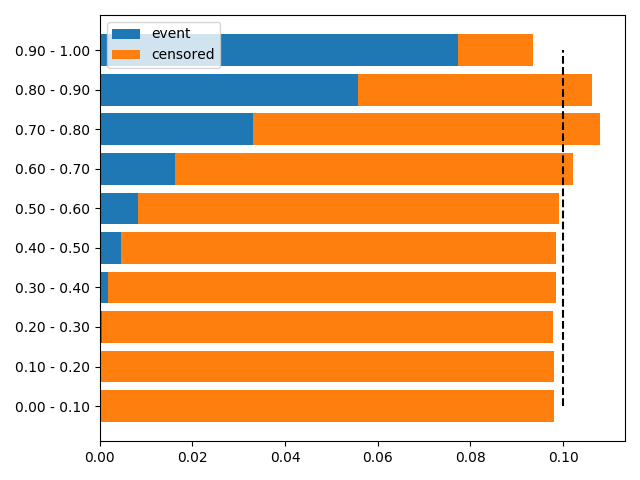}}
\end{figure}
\begin{figure}[htbp]
\floatconts
  {fig:auc_time}
  {\caption{The resulting AUC of eCaReNet over time. In the time range from 6 to 26 months after prostatectomy, the AUC is higher than 0.8. Overall, eCaReNet performs very similar to the expert pathologist. Only in the first months, where eCaReNet's survival curves are close to 1, it is outperformed by the pathologist, whose predictions are constant over time.}}
  {\includegraphics[width=\linewidth]{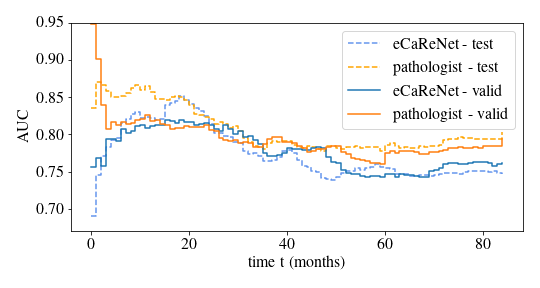}
  }
\end{figure}

\section{D-calibration}\label{apd:dcal}
In a d-calibrated model, the survival functions per patient can be interpreted as probability of relapse over time. If a survival curve shows 90\% survival probability, the patient can trust that only 10\% of patients with the same diagnosis experience a relapse at that time point. This also means that 10\% of patients should experience their event when the survival probability is between 90 and 100\%. Since the same holds for all other intervals (0-10, 10-20, ...), the expected number of events is compared to the true number of events and a chi-square test is used to measure this. Censored patients need to be treated differently from uncensored patients, because their true event time is not known. For details, see \citep{Haider2020}.

\section{Risk stratification details}
\label{apd:risk_strat}
For each patient, an individual risk score is calculated with \autoref{eq:risk}. It follows that $r\!=\!0$ if $\forall t_j\!:\!S(t_j)\!=\!1$ and $r\!=\!1$ if $\forall t_j\!:\!S(t_j)\!=\!0$. Risk scores are grouped into classes to enable a relative ranking among patients.
In order to assign risk scores to risk groups, intervals need to be defined, for which an exploratory approach is used. 

For the selection of the interval limits, multiple possible interval limits are defined and the best combination is chosen as follows. For each possible combination, the patients are assigned to the risk groups and patients within one group are combined in a single Kaplan-Meier curve \citep{Kaplan1958}. The resulting Kaplan-Meier curves per risk group are tested for discrimination power with a log-rank test, which is commonly used in survival analysis \citep{Li2015}. Since the proposed model allows for nonproportional hazards and therefore crossing survival curves, the log-rank test is modified with Fleming-Harrington weights according to \cite{Fleming2005}. If the test passes, the survival curves stratify well. 
Multiple combinations of boundaries can give perfectly stratified curves on the training set, which is why the best suited limits are further evaluated on the validation set. The limits with the best results on the validation set are used for final evaluation on the test set. The number of risk groups is also varied in this procedure, since using too few risk groups gives good stratification but is not meaningful for patients, whereas using too many risk groups, the Kaplan-Meier curves cannot separate well any more. In our analysis, using 8 risk groups was the largest number of possible groups that led to the best possible stratification in the training set. The found interval limits are 0.06, 0.12, 0.15, 0.18, 0.3, 0.42 and 0.51. Resulting Kaplan-Meier curves on the test set are shown in \figureref{fig:risk_stratification}. While all groups separate well in the training set, one log-rank test fails in the validation set and two tests fail for the test set. As limit for the p-value, 0.05 is chosen. 

\section{Attention example}
As an additional example for the experiment described in \sectionref{sec:attention}, \figureref{fig:apd_att} is provided. Also here it is shown that the malignant patches (in the lower part of the image) receive higher attention weights.
\begin{figure}[h]
\floatconts
{fig:apd_att}
{\caption{Another example of attention weights. The lower part is annotated as malignant (Gleason 5), the upper part as benign. }}
{\includegraphics[width=0.6\linewidth]{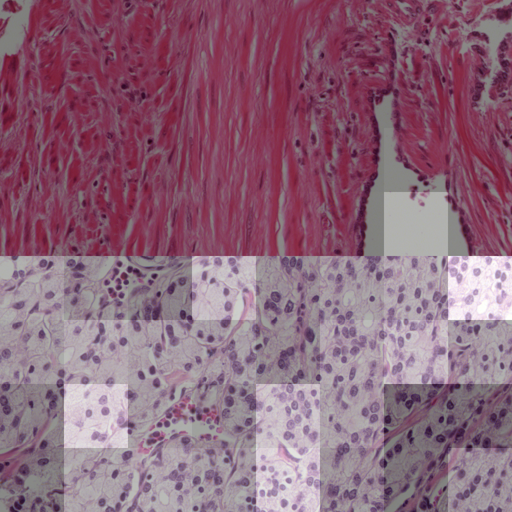}}
\end{figure}
\end{document}